\documentclass[10pt,twocolumn,letterpaper]{article}
\pdfoutput=1
\usepackage{cvpr}
\usepackage{times}
\usepackage{epsfig}
\usepackage{graphicx}
\usepackage{amsmath}
\usepackage{amssymb}
\usepackage{xspace}

\makeatletter
\DeclareRobustCommand\onedot{\futurelet\@let@token\@onedot}
\def\@onedot{\ifx\@let@token.\else.\null\fi\xspace}
\def\eg{\emph{e.g}\onedot} 
\def\ie{\emph{i.e}\onedot}

\def\wrt{w.r.t\onedot} 
\def\etal{\emph{et al}\onedot}
\makeatother

\usepackage[breaklinks=true,bookmarks=false]{hyperref}

\cvprfinalcopy

\newcommand{\PAR}[1]{\vskip4pt \noindent {\bf #1~}}
\newcommand{\PARbegin}[1]{\noindent {\bf #1~}}

\begin{document}
\newcommand{\partialderiv}[1]{\frac{\partial}{\partial #1}}
\newcommand{\partialfrac}[2]{\frac{\partial #1}{\partial #2}}
\newcommand{\deriv}[1]{\frac{\textnormal{d}}{\textnormal{d} #1}}
\newcommand{\derivfrac}[2]{\frac{\textnormal{d} #1}{\textnormal{d} #2}}
\newcommand{\half}{\nicefrac{1}{2}}
\newcommand{\veclen}[1]{\left\|#1\right\|}
\newcommand{\veccoords}[2]{
    \begin{pmatrix}
      #1 \\
      #2
    \end{pmatrix}
  }
  
\newcommand{\veccoordsthree}[3]{
    \begin{pmatrix}
      #1 \\
      #2 \\
      #3
    \end{pmatrix}
  }
  
\newcommand{\horizveccoords}[2]{
    \begin{pmatrix}
      #1 & #2
    \end{pmatrix}
  }

\newcommand{\floor}[1]{\lfloor #1\rfloor}  
\newcommand{\condprob}[2]{P\big(#1 \big| #2\big)}  
\newcommand{\prob}[1]{P\big(#1\big)}  

\newcommand{\at}[2]{\left.#1\right|_{#2}}
\newcommand{\newln}{\\&\quad\quad{}}
\newcommand{\tr}{^\top}
\newcommand{\inv}{^{-1}}
\newcommand{\reals}{\mathbb{R}}

\newcommand*\SetCond[2]{\left\{#1 \big| #2\right\}}
\newcommand*\Set[1]{\left\{#1\right\}}
\newcommand*\equ[1]{\begin{equation*}#1\end{equation*}}
\newcommand*\abs[1]{\left|#1\right|}
\newcommand*\range[2]{\Set{#1,\ldots,#2}}
\newcommand*\varsupto[2]{#1_1,\ldots,#1_{#2}}
\newcommand*\upto[1]{\in\range{1}{#1}}

\graphicspath{{images/}}

\title{Synthetic Occlusion Augmentation with Volumetric Heatmaps for the  \\ 2018 ECCV PoseTrack Challenge on 3D Human Pose Estimation}

\author{Istv\'{a}n S\'{a}r\'{a}ndi$^{1}$, Timm Linder$^{2}$, Kai O. Arras$^{2}$ and Bastian Leibe$^{1}$\vspace{8pt}\\
$^{1}$Visual Computing Institute, RWTH Aachen University \\ {\tt\small \{sarandi,leibe\}@vision.rwth-aachen.de} \\
$^{2}$Robert Bosch GmbH, Corporate Research \\ {\tt\small \{timm.linder,kaioliver.arras\}@de.bosch.com}%
}

\maketitle
\begin{abstract}
In this paper we present our winning entry at the 2018 ECCV PoseTrack Challenge on 3D human pose estimation. Using a fully-convolutional backbone architecture, we obtain volumetric heatmaps per body joint, which we convert to coordinates using soft-argmax. Absolute person center depth is estimated by a 1D heatmap prediction head. The coordinates are back-projected to 3D camera space, where we minimize the L1 loss. Key to our good results is the training data augmentation with randomly placed occluders from the Pascal VOC dataset. In addition to reaching first place in the Challenge, our method also surpasses the state-of-the-art on the full Human3.6M benchmark when considering methods that use no extra pose datasets in training. Code for applying synthetic occlusions is availabe at \url{https://github.com/isarandi/synthetic-occlusion}.
\end{abstract}

%%%%%%%%%%%%%%%%%%%%%%%%%%%%%%%%%%%%%%%%%%%%%%%%%%%%%%%%%%%%%%%%%%%%%%%%%%%%%%%%
\section{Introduction}

The 3D part of the 2018 ECCV PoseTrack Challenge invited participants to tackle the following task. Given an uncropped, static RGB image containing a single person, estimate the position of $J$ body joints in 3D camera space, relative to the root (pelvis) joint position.
Predicting human poses in 3D space has several important applications, such as human-robot collaboration and virtual reality. 

%%%%%%%%%%%%%%%%%%%%%%%%%%%%%%%%%%%%%%%%%%%%%%%%%%%%%%%%%%%%%%%%%%%%%%%%%%%%%%%%
\section{Related Work}

\PARbegin{3D Human Pose Estimation.} State-of-the-art 3D pose estimation methods are based on deep convolutional neural networks. We recommend Sarafianos \etal's survey~\cite{Sarafianos16CVIU} for an overview of methods. Recently, building on experience gained from 2D human pose estimation (\eg, \cite{Newell16ECCV}), heatmap-like methods have been introduced for 3D pose estimation with promising results. This includes volumetric heatmaps~\cite{Pavlakos17CVPR}\cite{Sun18ECCV}\cite{Luvizon18CVPR}, marginal heatmaps~\cite{Nibali18arXiv2} and location maps \cite{Mehta17TOG}.

\begin{figure}[tpb]
\centering
\includegraphics[scale=0.3]{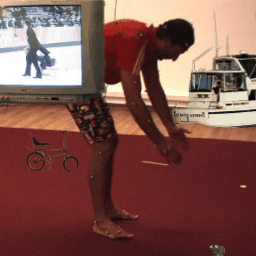} \hspace{0.5mm}
\includegraphics[scale=0.3]{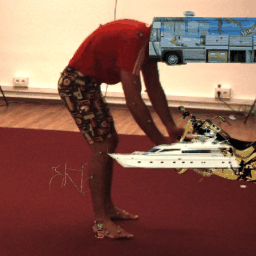} \vspace{1.5mm} \\
\includegraphics[scale=0.3]{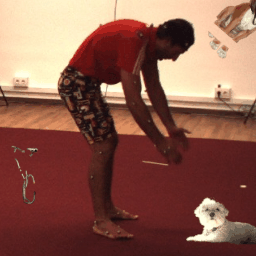} \hspace{0.5mm}
\includegraphics[scale=0.3]{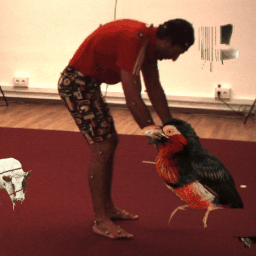}
\caption{Examples of synthetic occlusions with Pascal VOC objects (geometric and color augmentations not depicted).}
\label{fig:occlusion_types}
\end{figure}
\begin{figure*}[t]
\centering
\includegraphics[width=150mm]{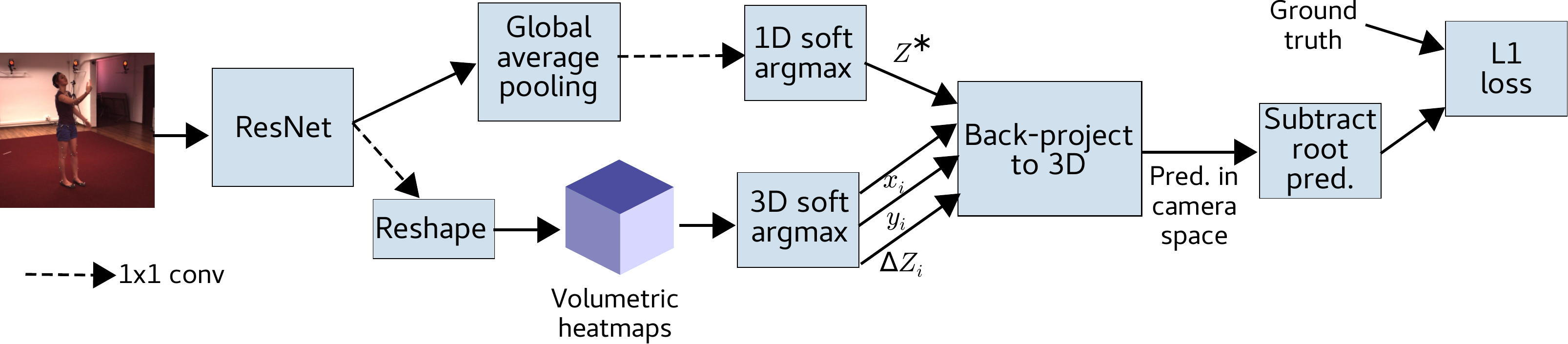}
\caption{Overview of our architecture.}
\label{fig:overview}
\end{figure*}

\PAR{Occlusion Augmentation.}
Erasing or pasting over parts of an image has been successfully used as data augmentation in image classification, object detection, person re-identification~\cite{Zhong17arXiv}\cite{DeVries17arXiv}\cite{Dwibedi17ICCV}\cite{Dvornik18ECCV}\cite{Georgakis17arXiv}, and facial facial landmark localization \cite{Yuen17TIV}. Ke \etal~\cite{Ke18arXiv} augment images for 2D pose estimation by copying background patches over some of the body joints. We have recently found that such techniques are also very effective for 3D pose estimation~\cite{Sarandi18IROSW}.

%%%%%%%%%%%%%%%%%%%%%%%%%%%%%%%%%%%%%%%%%%%%%%%%%%%%%%%%%%%%%%%%%%%%%%%%%%%%%%%%
\section{Dataset}
\label{sec:dataset}
The dataset in this challenge is a subset of Human3.6M~\cite{Ionescu14PAMI}\cite{Ionescu11ICCV}, with 35,832 training, 19,312 validation and 24,416 test examples. There are a few important differences compared to the full benchmark. First, the Challenge version lacks person bounding boxes and camera intrinsics as input. Second, the ground truth labels are more restricted, consisting only of camera-space 3D joint coordinates after subtraction of the root joint. Image-space joint coordinates are not available either.

%%%%%%%%%%%%%%%%%%%%%%%%%%%%%%%%%%%%%%%%%%%%%%%%%%%%%%%%%%%%%%%%%%%%%%%%%%%%%%%%
\section{Approach}
\label{sec:approach}
We present a modified version of the method we recently used for studying occlusion-robustness in 3D pose estimation~\cite{Sarandi18IROSW}, extending it to handle the above-mentioned differences.

\setlength\tabcolsep{1.9mm}
\begin{table*}[t]
\footnotesize
\centering
\begin{tabular}{lrrrrrrrrrrrrrrr|r}
\hline
                                  & Direct & Discuss    & Eat & Greet & Phone & Pose  & Purch. & Sit   & SitD  & Smoke & Photo & Wait  & Walk & WalkD &WalkT & Avg \\
\hline
% Rank   & A1 & A2 & A3 & A4 & A5 & A6 & A7 & A8 & A9  & A10 & A11 & A12 & A13 & A14 & A15 & Avg \\
Zhu      & 58 & 59 & 64 & 62 & 65 & 60 & 68 & 77 & 92  & 65  & 68  & 62  & 60  & 70  & 59  & 66  \\
Rhodin   & 51 & 53 & 58 & 52 & 64 & 53 & 67 & 94 & 132 & 65  & 64  & 57  & 53  & 67  & 53  & 66  \\
Zhou     & 52 & 56 & 55 & 51 & 57 & 53 & 64 & 73 & 81  & 61  & 60  & 57  & 49  & 61  & 53  & 59  \\
Park     & 53 & 52 & 52 & 53 & 55 & 55 & 54 & 71 & 84  & 56  & 60  & 58  & 51  & 64  & 57  & 58  \\
Shen     & 53 & 54 & 54 & 52 & 56 & 55 & 58 & 70 & 78  & 60  & 59  & 57  & 48  & 61  & 56  & 58  \\
Pavlakos & 44 & 46 & 50 & 47 & 56 & 47 & 52 & 63 & 70  & 54  & 54  & 48  & 46  & 58  & 46  & 52  \\
Sun      & \textbf{38} & 43 & 46 & 41 & 46  & \textbf{40} & 49 & 65 & 73  & 48 & 49  & 43  & 38  & 52  & \textbf{38}  & 47  \\
\hline
\textbf{Ours}     & \textbf{38} & \textbf{40} & \textbf{43} & \textbf{40} & \textbf{43} & \textbf{40} & \textbf{47} & \textbf{58} & \textbf{64}  & \textbf{43}  &  \textbf{48} & \textbf{42}  & \textbf{36}  & \textbf{50}  & \textbf{38}  & \textbf{45}  \\
\hline \\
\end{tabular}
\caption{Mean per joint position errors achieved by participants of the 2018 ECCV PoseTrack Challenge on 3D human pose estimation \cite{PoseTrackChallengeRankingsECCVW18}, on a subset of the Human3.6M dataset. In contrast to our method, some participants used extra 2D pose datasets in training (in accorance with the challenge rules).}
\label{tab:challenge_results}
\end{table*}

\setlength\tabcolsep{0.75mm}
\begin{table*}[t]
\footnotesize
\centering
\begin{tabular}{lrrrrrrrrrrrrrrr|r}
\hline
                               & Direct& Discuss&Eat  & Greet & Phone & Pose  & Purch.& Sit   & SitD  & Smoke & Photo & Wait  & Walk & WalkD &WalkT & Avg \\
\hline                                                                                                                
* Zhou (2017) \cite{Zhou17ICCV} & 54.8 & 60.7 & 58.2 & 71.4 & 62.0 & 65.5 & 53.8 & \textbf{55.6} & 75.2 & 111.6 & 64.2 & 66.0 & 51.4 & 63.2 & 55.3 & 64.9 \\
* Martinez (2017) \cite{Martinez17ICCV} & 51.8 & 56.2 & 58.1 & 59.0 & 69.5 & 55.2 & 58.1 & 74.0 & 94.6 & 62.3 & 78.4 & 59.1 & 65.1 & 49.5 & 52.4 & 62.9 \\
* Sun (2017) \cite{Sun17ICCV}           & 52.8 & 54.8 & 54.2 & 54.3 & 61.8 & 53.1 & 53.6 & 71.7 & 86.7 & 61.5 & 67.2 & 53.4 & 47.1 & 61.6 & 53.4 & 59.1 \\
* Pavlakos (2018) \cite{Pavlakos18CVPR} & 48.5 & 54.4 & 54.4 & 52.0 & 59.4 & 49.9 & 52.9 & 65.8 & 71.1 & 56.6 & 65.3 & 52.9 & 60.9 & \textbf{44.7} & 47.8 & 56.2 \\
* Luvizon (single-crop, 2018) \cite{Luvizon18CVPR} & 51.5 & 53.4 & 49.0 & 52.5 & 53.9 & 50.3 & 54.4 & 63.6 & 73.5 & 55.3 & 61.9 & 50.1 & 46.0 & 60.2 & 51.0 & 55.1 \\
* Luvizon (multi-crop, 2018) \cite{Luvizon18CVPR} & 49.2 & 51.6 & \textbf{47.6} & 50.5 & 51.8 & 48.5 & 51.7 & 61.5 & 70.9 & 53.7 & 60.3 & 48.9 & 44.4 & 57.9 & 48.9 & 53.2 \\
* Sun (2018) \cite{Sun18ECCV} & \textbf{47.5} & \textbf{47.7} & 49.5 & \textbf{50.2} & \textbf{51.4} & \textbf{43.8} & \textbf{46.4} & 58.9 & \textbf{65.7} & \textbf{49.4} & \textbf{55.8} & \textbf{47.8} & \textbf{38.9} & 49.0 & \textbf{43.8} & \textbf{49.6} \\
\hline
Tekin (2016) \cite{Tekin16CVPR}       & 102.4 & 147.7 & 88.8 & 125.4 & 118.0 & 112.4 & 129.2 & 138.9 & 224.9 & 118.4 & 182.7 & 138.8 & 55.1 & 126.3 & 65.8 & 125.0 \\
Zhou (2016) \cite{Zhou16CVPR}         & 87.4  & 109.3 & 87.1 & 103.2 & 116.2 & 106.9 & 99.8  & 124.5 & 199.2 & 107.4 & 139.5 & 118.1 & 79.4 & 114.2 & 97.7 & 113.0 \\
Xingyi (2016) \cite{Zhou16ECCV}       & 91.8  & 102.4 & 97.0 & 98.8  & 113.4 & 90.0  & 93.8  & 132.2 & 159.0 & 106.9 & 125.2 & 94.4  & 79.0 & 126.0 & 99.0 & 107.3 \\
Sun (2017) \cite{Sun17ICCV}           & 90.2  & 95.5  & 82.3 & 85.0  &  87.1 & 87.9  & 93.4  & 100.3 & 135.4 & 91.4  & 94.5  & 87.3  & 78.0 & 90.4  & 86.5 & 92.4  \\
Pavlakos (2017) \cite{Pavlakos17CVPR} & 67.4  & 72.0  & 66.7 & 69.1  &  72.0 & 65.0  & 68.3  &  83.7 &  96.5 & 71.7  & 77.0  &  65.8 & 59.1 & 74.9  & 63.2 & 71.9 \\
Sun (2018) \cite{Sun18ECCV}           & 63.8  & 64.0  & 56.9 & 64.8  &  62.1 & 59.8  & 60.1  &  71.6 &  91.7 & 60.9  & 70.4  &  65.1 & 51.3 & 63.2  & 55.4 & 64.1  \\
\hline 
\textbf{Ours (no occlusion augmentation)}          & 63.3 & 65.5 & 56.0 & 62.1 & 64.0 & 60.7 & 64.8 & 76.7 & 93.0 & 63.3 & 69.7 &  62.0 & 54.1 & 68.8 & 61.3 & 65.7 \\
\textbf{Ours (full)} & \textit{49.1} & \textit{54.6} & \textit{50.4} & \textit{50.7} & \textit{54.8} & \textit{47.4} & \textit{50.1} & \textit{67.5} & \textit{78.4} & \textit{53.1} & \textit{57.4} & \textit{50.7} & \textit{40.1} & \textit{54.0} & \textit{46.1} & \textit{54.2} \\
\hline \\
\end{tabular}
\caption{Mean per joint position error on the full Human3.6M dataset. Results marked with an asterisk (*) were achieved using extra 2D pose dataset(s) in training. Boldface indicates the overall best results, while italic indicates the best when using no extra 2D pose datasets.}
\label{tab:comparison_prior_work}
\end{table*}

\PAR{Image Preprocessing.}
We obtain person bounding boxes using the YOLOv3 detector~\cite{Redmon18arXiv}. Treating the original camera's focal length $f$ as a global hyperparameter, we reproject the image to be centered on the person box, at a scale where the larger side of the box fills 90\% of the output.

\PAR{Backbone Network.}
We feed the cropped and zoomed image ($256 \times 256$~px) into a fully-convolutional backbone network (ResNet~v2-50~\cite{He16ECCV}\cite{Silberman17Github}). We directly obtain volumetric heatmaps from the backbone net by adding a 1x1 convolutional layer on the last spatial feature map of the backbone, producing $J\cdot D$ output channels. The resulting tensor is reshaped to yield $J$ volumes, one per body joint, each with depth $D$.

\PAR{Volumetric Heatmaps.} We follow Pavlakos \etal in the interpretation of the volumetric heatmap's axes~\cite{Pavlakos17CVPR}: X and Y correspond to image space and the depth axis to camera space, relative to the person center. Relative depths are not sufficient, however, when back-projecting from image to camera space. Pavlakos \etal optimize the root joint depth in post-processing, based on bone-length priors. By contrast, we predict it using a second prediction head on the backbone net (see Fig. \ref{fig:overview}). This outputs a 1D heatmap discretized to 32 units, representing a 10 meter range in front of the camera.

\PAR{Soft-Argmax.}
We extract coordinate predictions from the heatmaps using soft-argmax~\cite{Levine16JMLR}\cite{Nibali18arXiv}. Since this operation is differentiable, there is no need to provide ground-truth heatmaps at training time~\cite{Sun18ECCV}. Instead, the loss can be computed deeper in the network and backpropagated through the soft-argmax operation. Soft-argmax also reduces the quantization errors inherent in hard argmax and gives fine-grained, continuous results without requiring memory-expensive, high-resolution heatmaps~\cite{Sun18ECCV}. Indeed, we use a heatmap resolution as low as $16^3$ for the results presented in this paper.

\PAR{Camera Intrinsics.} Having predicted image coordinates $x_i$,~$y_i$, depth coordinates $\Delta Z_i$ relative to the person center and the absolute depth $Z^*$ of the person center by soft-argmax, we now need camera intrinsics to move from image space to 3D camera space. As mentioned earlier, the original camera's focal length $f$ is treated as a hyperparameter, and we must also take into account the zooming factor $s$ applied in preprocessing.

To avoid the need for precise hyperparameter tuning of $f$, we learn an additional, input-independent corrective factor $c$ for the focal length during training, to achieve better alignment of image and heatmap locations. Denoting the image height and width as $H$ and $W$, back-projection is performed as
\begin{equation}
\veccoordsthree{X_i}{Y_i}{Z_i} \hspace{-2pt}= \hspace{-2pt}(Z^*+\Delta Z_i)\hspace{-2pt}
\begin{pmatrix}
fsc & 0 & W/2 \\
0 & fsc & H/2 \\
0 & 0 & 1
\end{pmatrix}\inv \hspace{-5pt} \veccoordsthree{x_i}{y_i}{1}.
\end{equation}
\PAR{Loss.}
After subtracting the root joint coordinates, we compute the $L^1$ loss in the original camera space \wrt the provided root-relative ground truth. No explicit heatmap loss is used. Since all above operations are differentiable the whole network can be trained end-to-end.

\PAR{Data Augmentation.}
In our recent study on the occlusion-robustness of 3D pose estimation~\cite{Sarandi18IROSW}, we found that augmenting training images with synthetic occlusions acts as an effective regularizer. Starting with the objects in the Pascal VOC dataset~\cite{Everingham12}, we filter out persons, segments labeled as \emph{difficult} or \emph{truncated} and segments with area below 500 px, leaving 2638 objects. With probability $p_{\text{occ}}$, we paste a random number (between 1 and 8) of these objects at random locations in each frame.
We also apply standard geometric augmentations (scaling, rotation, translation, horizontal flip) and appearance distortions (blurs and color manipulations). At test time only horizontal flipping augmentation is used.

\PAR{Training Details.}
The backbone net is initialized with ImageNet-pretrained weights from~\cite{Silberman17Github}. We train the final method for 410 epochs on the union of the training and validation set using the Adam optimizer and cyclical (triangular) learning rates~\cite{Smith17WACV}. Our final challenge predictions were produced using a snapshot ensemble~\cite{Huang17arXiv}, averaging the predictions of snapshots taken at the last three learning-rate-minima of the cyclical schedule. We used $f=1500$ and $p_{\text{occ}}=0.5$ for the submission.

%%%%%%%%%%%%%%%%%%%%%%%%%%%%%%%%%%%%%%%%%%%%%%%%%%%%%%%%%%%%%%%%%%%%%%%%%%%%%%%%
\section{Results}
The evaluation metric is the mean per joint position error (MPJPE) over all joints after subtraction of the root joint position. Our method achieves best results for all actions, even ahead of methods using extra 2D pose datasets in training (see Table \ref{tab:challenge_results}). The margin is largest for the actions \emph{Sitting} and \emph{Sitting Down}, showing that our method is more robust to the presence of a chair, which is the only occluding object in the Human3.6M dataset.

\begin{figure}[tpb]
\centering
\includegraphics[width=83mm]{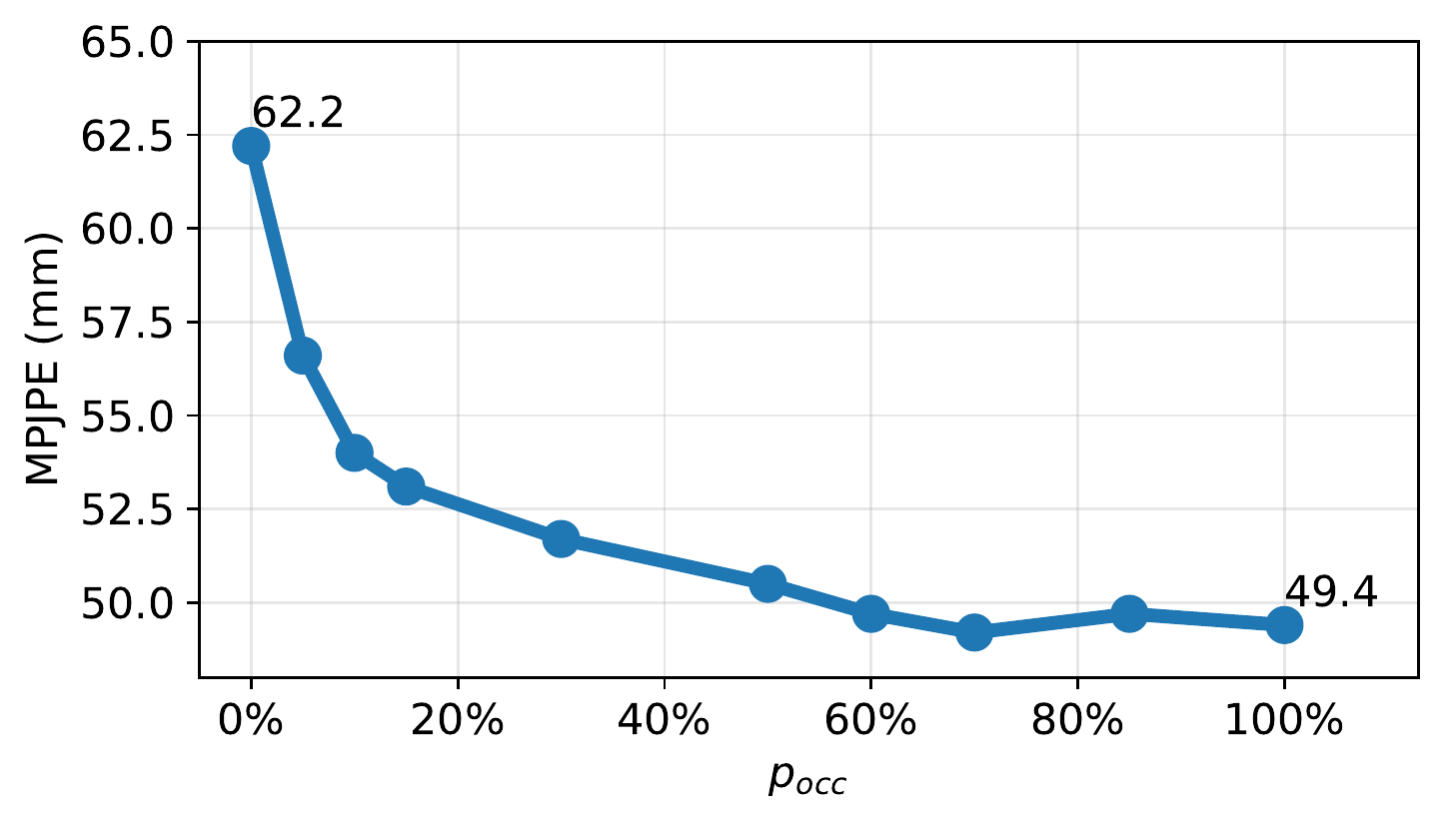}
\caption{Effect of the per-frame probability of occlusion augmentation ($p_{occ}$) evaluated on the Challenge validation set.}
\label{fig:pocc}
\end{figure}

\PAR{Effect of Occlusion Augmentation.}
Fig. \ref{fig:pocc} shows how synthetic occlusion augmentation improves results on the Challenge validation set as we vary the probability $p_{\text{occ}}$ of applying occlusion augmentation to each frame. Augmenting just 10\% of the images already improves MPJPE by 8.2~mm and improvements continue to about $p_{\text{occ}}=70\%$, after which performance is only influenced slightly.

\vspace{2mm}

\PAR{Full Human3.6M Benchmark.}
For comparison with prior work, we train and evaluate our method on the full Human3.6M benchmark as well (Table \ref{tab:comparison_prior_work}). Here we use the bounding boxes and camera intrinsics provided with the dataset and minimize the $L^1$ loss computed on the absolute (\ie non-root-relative) coordinates in camera space for 40 epochs. The person center depth $Z^*$ is estimated as described in Section \ref{sec:approach}. We follow the common protocol of training on five subjects (S1, S5, S6 S7, S8) and evaluating on two (S9, S11), without Procrustes alignment. We use no snapshot ensembling here, for better comparability. The occlusion probability $p_{occ}$ is set to 1. Our method outperforms all prior work on Human3.6M when no additional pose datasets are used for training.

%%%%%%%%%%%%%%%%%%%%%%%%%%%%%%%%%%%%%%%%%%%%%%%%%%%%%%%%%%%%%%%%%%%%%%%%%%%%%%%%
\section{Conclusion}

We have presented an architecture and data augmentation method for 3D human pose estimation and have shown that it outperforms other methods both by achieving first place in the 2018 ECCV PoseTrack Challenge and by surpassing the state-of-the-art on the full benchmark among methods using no additional pose datasets in training.

%%%%%%%%%%%%%%%%%%%%%%%%%%%%%%%%%%%%%%%%%%%%%%%%%%%%%%%%%%%%%%%%%%%%%%%%%%%%%%%%
\PAR{Acknowledgments.} This project has been funded by a grant from the Bosch Research Foundation, by ILIAD (H2020-ICT-2016-732737) and by ERC Consolidator Grant DeeViSe (ERC-2017-COG-773161).

{\small
\bibliographystyle{ieee}
\bibliography{abbrev_short,references}
}

\end{document}